\documentclass[letterpaper, 10 pt, conference]{ieeeconf}  

\IEEEoverridecommandlockouts                              

\usepackage{graphics} 
\usepackage{epsfig} 

\usepackage{amsmath} 
\usepackage{amssymb}  
\usepackage{amsthm}
\usepackage{subcaption}
\usepackage{float}
\usepackage{todonotes}
\usepackage{neuralnetwork}
\usepackage[ruled,linesnumbered]{algorithm2e}
\usepackage{soul}
\let\labelindent\relax
\usepackage{enumerate}
\usepackage{enumitem}
\usepackage{changepage} 

\newlist{steps}{enumerate}{1}
\setlist[steps, 1]{label = \textbf{Step \arabic*)}}

\newtheorem{example}{Example}

\DeclareMathOperator{\arctantwo}{arctan2}

\DeclareMathAlphabet{\pazocal}{OMS}{zplm}{m}{n}
\title{\LARGE \bf
Hysteresis-Based RL: Robustifying Reinforcement Learning-based Control Policies via Hybrid Control}

\author{Jan de Priester$^{1}$, Ricardo G. Sanfelice$^{2}$, and Nathan van de Wouw$^{3}$
\thanks{*Research by R. G. Sanfelice has been partially supported by the National Science Foundation under Grant no. ECS-1710621, Grant no. CNS-1544396, and Grant no. CNS-2039054, by the Air Force Office of Scientific Research under Grant no. FA9550-19-1-0053, Grant no. FA9550-19-1-0169, and Grant no. FA9550-20-1-0238, and by the Army Research Office under Grant no. W911NF-20-1-0253.}
\thanks{$^{1}${\tt\small jandepriester@gmail.com}}
\thanks{$^{2}$Ricardo G. Sanfelice is with the Department of Electrical and Computer Engineering, University of California, Santa Cruz, CA 95064, USA;
        {\tt\small ricardo@ucsc.edu}}%
\thanks{$^{3}$Nathan van de wouw is with the Department of Mechanical Engineering, Eindhoven University of Technology, Eindhoven, 5612 AZ, Netherlands;
        {\tt\small n.v.d.wouw@tue.nl}}%
}

\AtBeginDocument{\colorlet{defaultcolor}{.}}
\begin{document}

\maketitle
\thispagestyle{empty}
\pagestyle{empty}


\begin{abstract}

Reinforcement learning (RL) is a promising approach for deriving control policies for complex systems. As we show in two control problems, the derived policies from using the Proximal Policy Optimization (PPO) and Deep Q-Network (DQN) algorithms may lack robustness guarantees. Motivated by these issues, we propose a new hybrid algorithm, which we call Hysteresis-Based RL (HyRL), augmenting an existing RL algorithm with hysteresis switching and two stages of learning. We illustrate its properties in two examples for which PPO and DQN fail.

\end{abstract}

\section{INTRODUCTION}
Reinforcement learning (RL) is an area of machine learning that focuses on how an agent (or controller) should act in an environment to maximize a cumulative reward. Over the years, work has been done on improving the robustness of the policy derived from RL algorithms. In~\cite{ltjens2019certified}, an approach is discussed to increase robustness to adversarial noise on sensory inputs for RL-based policies. In~\cite{Tessler2019ActionRobust_disttoactiontaken}, robustness of RL policies is treated for two scenarios, namely, when an alternative action is taken and when a perturbation is added to the selected action. In~\cite{Mankowitz2020Robust}, a framework is discussed to incorporate robustness against perturbations in the transition dynamics. In~\cite{Zhang2021RobustRLAdversary}, the robustness of reinforcement learning is studied with adversarially perturbed state observations. However, for problems suffering from topological obstructions, such as the problem of stabilizing a disjoint set or the problem of globally steering to a target with obstacle avoidance, the policies resulting from RL methods are often not robust to small measurement noise, as we illustrate in this paper.

To illustrate that RL methods yield policies that are not robust, consider the problem of steering an autonomous vehicle to move past an obstacle so as to reach a target.
\begin{figure}[b]
    \centering
    \includegraphics[width = 1\linewidth]{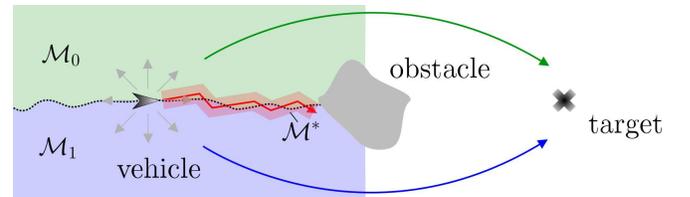}
    \caption{Trajectories of the autonomous vehicle example. The vehicle is depicted by the gray arrow, the obstacle by the gray block, and the target by the cross. The green and blue arrows represent the trajectories obtained with the policy derived from the used RL method when the vehicle is in $\pazocal{M}_0$ and $\pazocal{M}_1$, respectively. The dotted line represents the set of critical points $\pazocal{M}^*$. The red arrow depicts the trajectory of the vehicle in the presence of small measurement noise. The red area represents the set of points the vehicle can observe along the trajectory due to small measurement noise.}
    \label{fig:intro:carcrashWithDist}
\end{figure}
After successfully training the autonomous vehicle, a policy is found that steers the vehicle either left or right past the obstacle to clear it, rendering a location past the obstacle (the target) asymptotically stable. In Figure~\ref{fig:intro:carcrashWithDist}, the policy and trajectory of the autonomous vehicle are shown in the presence of small measurement noise. The policy is simple: the vehicle steers left when in the green area ($\pazocal{M}_0$) and right when in the blue area ($\pazocal{M}_1$). However, when the vehicle is near the boundary between the two areas ($\pazocal{M}^*$), the vehicle can physically be in $\pazocal{M}_0$ but, due to noise in the measurements of position, perceive itself to be in $\pazocal{M}_1$. As a result, the vehicle turns right instead of left. Vice versa, the vehicle can physically be in $\pazocal{M}_1$ but perceive itself to be in $\pazocal{M}_0$ and turn left instead of right. Repetition of this occurrence causes the vehicle to drive straight into the obstacle, as shown in Figure~\ref{fig:intro:carcrashWithDist}. This behavior is unfavorable because, for a small amount of measurement noise, the vehicle can crash into the obstacle with such a policy, making it not robust against small measurement noise. In fact, such unfavorable behavior can arise under arbitrarily small measurement noise.

Motivated by this issue, we propose a new hybrid RL algorithm, which we call Hysteresis-Based RL (HyRL). The algorithm is designed to prevent solutions from getting stuck around critical points in the presence of small measurement noise. The set of points from which solutions evolve in opposite directions is referred to as the set of critical points, e.g., the set $\pazocal{M}^*$ in Figure~\ref{fig:intro:carcrashWithDist}. HyRL implements a hysteresis effect with the use of a hybrid system, see Section~\ref{sec:HybridDQN}. In Section~\ref{sec:motivation}, we present two examples with critical points where the resulting policies from Proximal Policy Optimization~(PPO)~\cite{PPO_Schulman_2017} and Deep Q-Network~(DQN)~\cite{DQNintro} are not robust to small measurement noise. The ideas in this paper can be naturally extended for other RL methods that have a similar policy parameterization, such as Soft Actor-Critic ~(SAC)~\cite{SACpaper}, Deep Deterministic Policy Gradient~(DDPG)~\cite{LillicrapHPHETS15DDPG}, and both Advantage Actor-Critic~(A2C) and Asynchronous Advantage Actor-Critic~(A3C)~\cite{mnih2016asynchronousA2CA3C}. The HyRL algorithm is employed to obtain a hybrid closed-loop system that is robust against small measurement noise. In Section~\ref{sec:hybridsims}, the two examples from Section~\ref{sec:motivation} are revisited to show that the hybrid closed-loop system obtained with the HyRL algorithm is indeed robust against small measurement noise. 

\section{PRELIMINARIES}\label{sec:prelims}
In this section, the used mathematical notation, Markov decision processes, the DQN and PPO algorithms, and hybrid systems are introduced. 
\subsection{Notation}
The following notation is used throughout the paper. The $n$-dimensional Euclidean space is denoted by~$\mathbb{R}^n$. The real numbers are denoted by~$\mathbb{R}$. The nonnegative real numbers are denoted by~$\mathbb{R}_{\geq0}$, i.e., $\mathbb{R}_{\geq0}:=[0,\infty)$. The natural numbers including~0 are denoted by $\mathbb{N}$, i.e., $\mathbb{N}:=\{0,1,2,...\}$. The open unit ball, of appropriate dimension and centered at the origin, in the Euclidean norm is denoted by~$\mathbb{B}$.
\subsection{Markov Decision Processes}
In this paper, Markov decision processes (MDPs) are used as a formalism for RL~\cite{mdpBook}. In an MDP, the learner/controller is referred to as the agent. The agent is deployed in an environment and interacts with it. These interactions continually occur. At each time step $k$, the agent receives some representation of the state of the environment $S_k \in \pazocal{S}$, where $\pazocal{S}$ is the state space, in the form of an observation $o_k(S_k)$, and uses it to select an action $A_k \in \pazocal{A}$, where $\pazocal{A}$ is the action space. At the next time step, the agent has transitioned into a new state $S_{k+1}$ as a result of $A_k$, and perceives a reward $R_{k+1} \in \pazocal{R} \subset \mathbb{R}$, where $\pazocal{R}$ is the reward space and $\mathbb{R}$ the Euclidean space. The goal of RL is to train an agent how to behave so as to maximize a numerical reward quantity. The numerical reward quantity is the expected (discounted) return 
\begin{equation}
    G_k:= \sum_{l=0}^\infty \gamma^l R_{k+l+1},
    \label{eq:discountedreturn}
\end{equation}
which is the (weighted) sum of all rewards the agent experiences during a training episode, where $\gamma \in [0,1]$ is the discount factor. As $\gamma$ approaches zero, the agent becomes myopic and is only concerned with maximizing immediate rewards. As $\gamma$ approaches one, the agent becomes far-sighted and takes into account the value of future rewards more heavily.

In this paper, the example problems are formulated as continuous-time problems with dynamics $\dot{\xi} = f(\xi, u)$, reward function $R(\xi)$, and observation function $o(\xi)$, where $\xi$ is the state and $u$ the control input, both parameterized by ordinary time~$t$. The equivalent system in an MDP formulation is obtained by substituting $\xi$ for $S_k$, $u$ for $A_k$, $R(\xi)$ for $R_{k+1}(S_{k+1})$, $o(\xi)$ for $o_{k}(S_k)$, where $k$ is the time step associated to $t$. The state-transition model is given by $S_{k+1} = S_k + f(S_k,A_k) \Delta t$, where $\Delta t$ is the sampling time.

\subsection{Reinforcement Learning Methods}
In this paper, two different RL methods are used, namely, PPO and DQN. PPO is used for the first example problem where the action space is continuous, and DQN is used for the second example problem where the action space is discrete.

PPO is a policy-based method, hence the goal is to approximate the stochastic policy~$\pi^s(u|o(\xi)):\pazocal{S}\times \pazocal{A} \rightarrow [0,1]$. A stochastic policy returns the probability of selecting action~$u$ given observation~$o(\xi)$. A control policy~$\pi^*(o(\xi)):\pazocal{S}\rightarrow \pazocal{A}$ is derived by sampling an action~$u$ from the stochastic policy given an observation~$o(\xi)$, that is~$\pi^*(o(\xi)) \sim \pi^*(u|o(\xi))$. The stochastic policy is parameterized by a multivariate Gaussian distribution given by
\begin{equation}
    \pi^s(u|o(\xi)) = \pazocal{N}(\mu(o(\xi)),\Sigma^2(o(\xi))),
\end{equation}
where $\mu(o(\xi))$ denotes the mean vector and $\Sigma(o(\xi))$ the covariance matrix of the multivariate Gaussian distribution~$\pazocal{N}$. Neural networks are used to estimate the mean vector and covariance matrix. 

DQN is an action-value-based method, hence the goal is to approximate the action-value function $Q(o(\xi),u)$. The action-value function is an estimate of the expected (discounted) return when starting in state $\xi$, taking action $u$, and following the target policy afterwards, that is, the policy used to update the weights of the network. The function can be seen as a measure of how beneficial it is to take an action $u$ at a state $\xi$. A control policy $\pi^*(o(\xi)):\pazocal{S}\rightarrow \pazocal{A}$ is derived from the action-value function by taking the action that maximizes the action-value function (greedy action), that is
\begin{equation}
    \pi^*(o(\xi)) = \underset{u\in \pazocal{A}}{\arg \max }\: Q(o(\xi),u).
    \label{eq:prelim:policy}
\end{equation}
A neural network is used to estimate the action-value function.
\subsection{Hybrid Systems}\label{sec:hybridsystems}
A hybrid system $\pazocal{H}=(C, F, D, G)$ is defined as
\begin{equation}
\pazocal{H}:\left\{\begin{array}{ll}{\dot{{z}}\phantom{^+}=F({z})} & {{z} \in C} \\ {{z}^{+}=G({z})} & {{z} \in D}\end{array}\right.,
\label{eq:prelim:hybridsystem}
\end{equation}
where ${z}~\in~\mathbb{R}^n$ denotes the state variable, ${z}^{+}$ the state variable after a jump, $F:C~\rightarrow~\mathbb{R}^n$ is a function referred to as the flow map, $C~\subset~\mathbb{R}^n$ is the set of points referred to as the flow set, $G:D~\rightarrow~\mathbb{R}^n$ the jump map, and $D~\subset~\mathbb{R}^n$ the jump set. When the state is in the flow set, the system evolves continuously and is described by the differential equation defined by the flow map. When the state is in the jump set, the state is updated using the difference equation defined by the jump map. In this way, with some abuse of notation, the solution to~(\ref{eq:prelim:hybridsystem}) is given by a function $(t,j) \mapsto {z}(t,j)$ defined on a hybrid time domain, which properly collects values of the ordinary time variable $t \in \mathbb{R}_{\geq0}$ and of the discrete jump variable $j \in \mathbb{N}$. The hybrid system $\pazocal{H}$ allows for the combination of continuous-time behavior~(flow) with discrete-time behavior~(jumps). For more details on hybrid dynamical systems, see \cite{ HybridFeedbackControl2021Book},\cite{hybriddynamical}.

\section{MOTIVATION}\label{sec:motivation}
In this section, two examples, one pertaining to robustly stabilizing a set-point on the unit circle (continuous action space) and another to robustly avoiding an obstacle (discrete action space), are presented where the resulting control policies from PPO and DQN are not robust to small measurement noise. These examples motivate the design of HyRL in Section~\ref{sec:HybridDQN}. For the unit circle example, a control policy is derived with the use of PPO\footnote{Code at {github.com/HybridSystemsLab/UnitCircleHyRL}} and for the obstacle avoidance example DQN\footnote{Code at {github.com/HybridSystemsLab/ObstacleAvoidanceHyRL}} is used. We utilize the implementations provided by the Stable Baselines3 library for the PPO and DQN algorithms~\cite{stable-baselines3}. For the unit circle problem, we show that PPO can lead the system to get stuck away from the set-point in the presence of small measurement noise. For the obstacle avoidance problem, we show DQN can cause the system to crash into the obstacle in the presence of small measurement noise.

\begin{example}[Continuous action space problem]\label{sec:unitcircleDQN}
The unit circle problem, as discussed in \cite{4739481}, is considered. The objective is to robustly globally asymptotically stabilize the set-point $\xi^* = [1$~$0]^\top$ for the constrained system
\begin{equation}
   \dot{\xi} = f(\xi,u) := u \left[\begin{array}{cc} 0 & -1 \\ 1 & 0 \end{array}\right] \xi \quad \xi = \left[ \begin{array}{c}
         x  \\
         y 
    \end{array}{}\right] \in \pazocal{S}^{1},
    \label{eq:UnitcircleDyn}
\end{equation}
where $u \in [-1,1]$ is the control input, $\xi$ the state vector, and $\pazocal{S}^1$ the unit circle defined as $\pazocal{S}^1 :=\left\{\xi \in \mathbb{R}^{2}\mid | \xi |=1\right\}$. This model describes the evolution of a point on a circle as a function of the control input $u$. For this problem the observation function $o(\xi)$ is equal to the state $\xi$ of the system, namely, $o(\xi) = \xi$. 

To globally stabilize the set-point $\xi^*$, a reward function is required that has a global maximum at the set-point $\xi = \xi^*$. One such reward function is given by
\begin{equation}
    R(x,y) = -\frac{1}{\pi}|\arctantwo(y,x) |  ,
    \label{eq:RewardUnitCircle}
\end{equation}
where $\arctantwo(y,x)$ is the four-quadrant inverse tangent of $(x,y)$. The angle is divided by $\pi$ to normalize the reward function such that $R(x,y) \in [-1, 0]$. The reward function has a global minimum at $\xi = [-1$~$0]^\top$, and is symmetric in the sense that $R(x,y) = R(x,-y)$. Therefore, it is expected that for a small change in $\xi$ around the point $\xi = [-1$~$0]^\top$, different control policies are found, e.g., $u = 1$ for $y>0$, and $u = -1$ for $y<0$. The stochastic policy $\pi^s(u|o(\xi))$ found numerically is visualized in the left figure in Figure~\ref{fig:policymapandsimulation_UC}.
In the left figure in Figure~\ref{fig:policymapandsimulation_UC}, it can be seen that the policy, roughly speaking, pushes the system clockwise for $y>0$ and counterclockwise for $y<0$. In the right figure in Figure~\ref{fig:policymapandsimulation_UC}, the system is simulated for initial conditions $\xi_0^1 = [-0.81$~$0.59]^\top$, $\xi_0^2 = [-0.95$~$-0.31]^\top$, and $\xi_0^3 = [-1$~$0]^\top$. For this simulation, a small amount of noise of magnitude $ 0.1$ is added to the measurement of the state $\xi$. In the right figure in Figure~\ref{fig:policymapandsimulation_UC}, it can be seen that the solution starting from $\xi_0^1$, i.e., $y>0$, moves clockwise to the set-point $\xi^*$ and the solution starting from $\xi_0^2$, i.e., $y<0$, moves counterclockwise to the set-point $\xi^*$. However, the solution starting from $\xi_0^3$ gets stuck nearby its starting point as a result of the small measurement noise. The found policy is not robust against small measurement noise, meaning that for a small amount of noise the system can get stuck around this point, as can be seen in Figure~\ref{fig:policymapandsimulation_UC}. The point~$\xi_c= [-1$~$0]^\top$ is considered a critical point for this problem. 

\begin{figure}[]
    \centering
\begin{subfigure}[t]{.5\linewidth}
    \centering
    \includegraphics[width = 1\linewidth, height=3.25cm]{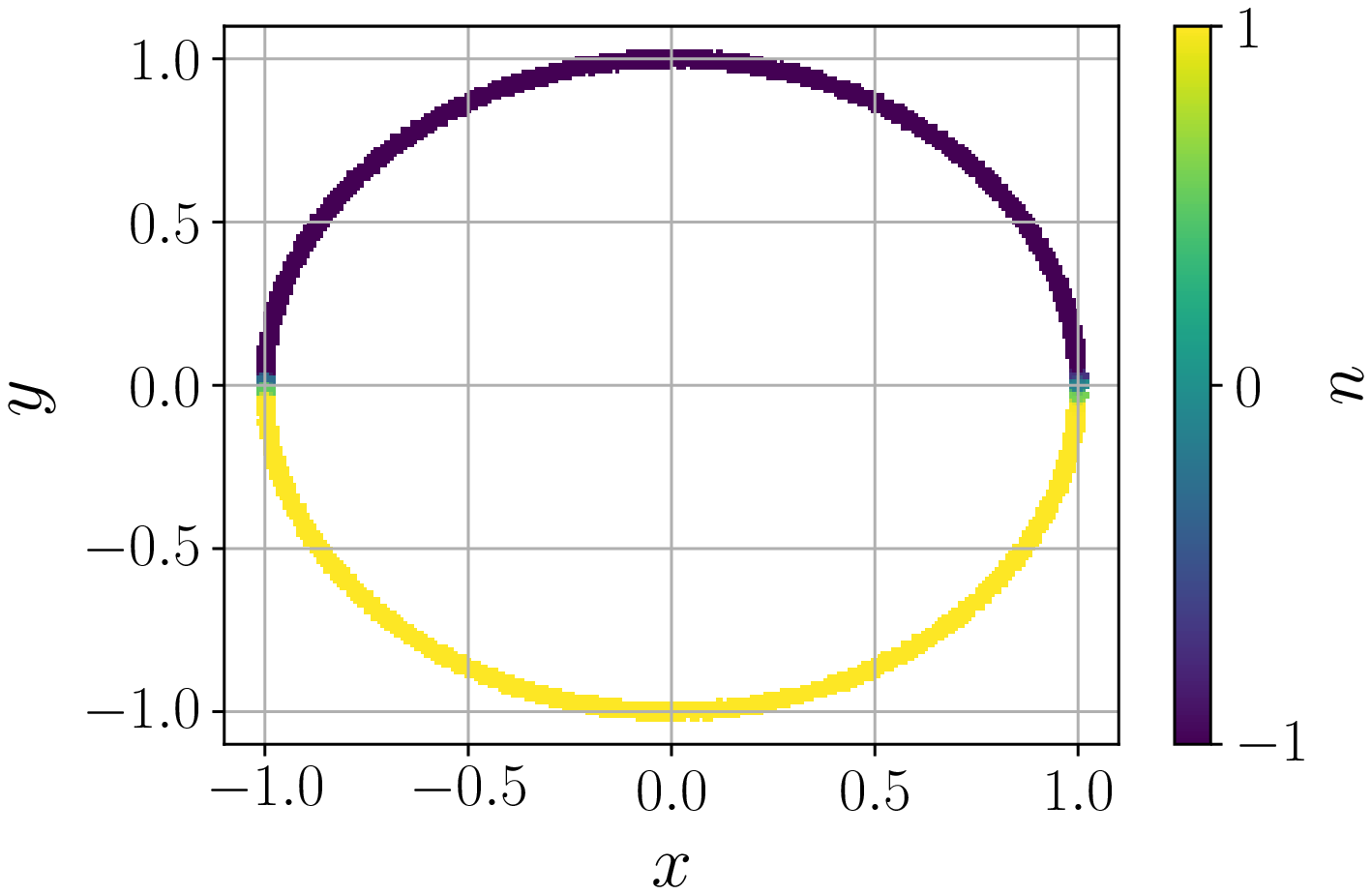}
\end{subfigure}%
\begin{subfigure}[t]{.5\linewidth}
    \centering
    \includegraphics[width = 1\linewidth, height=3.25cm]{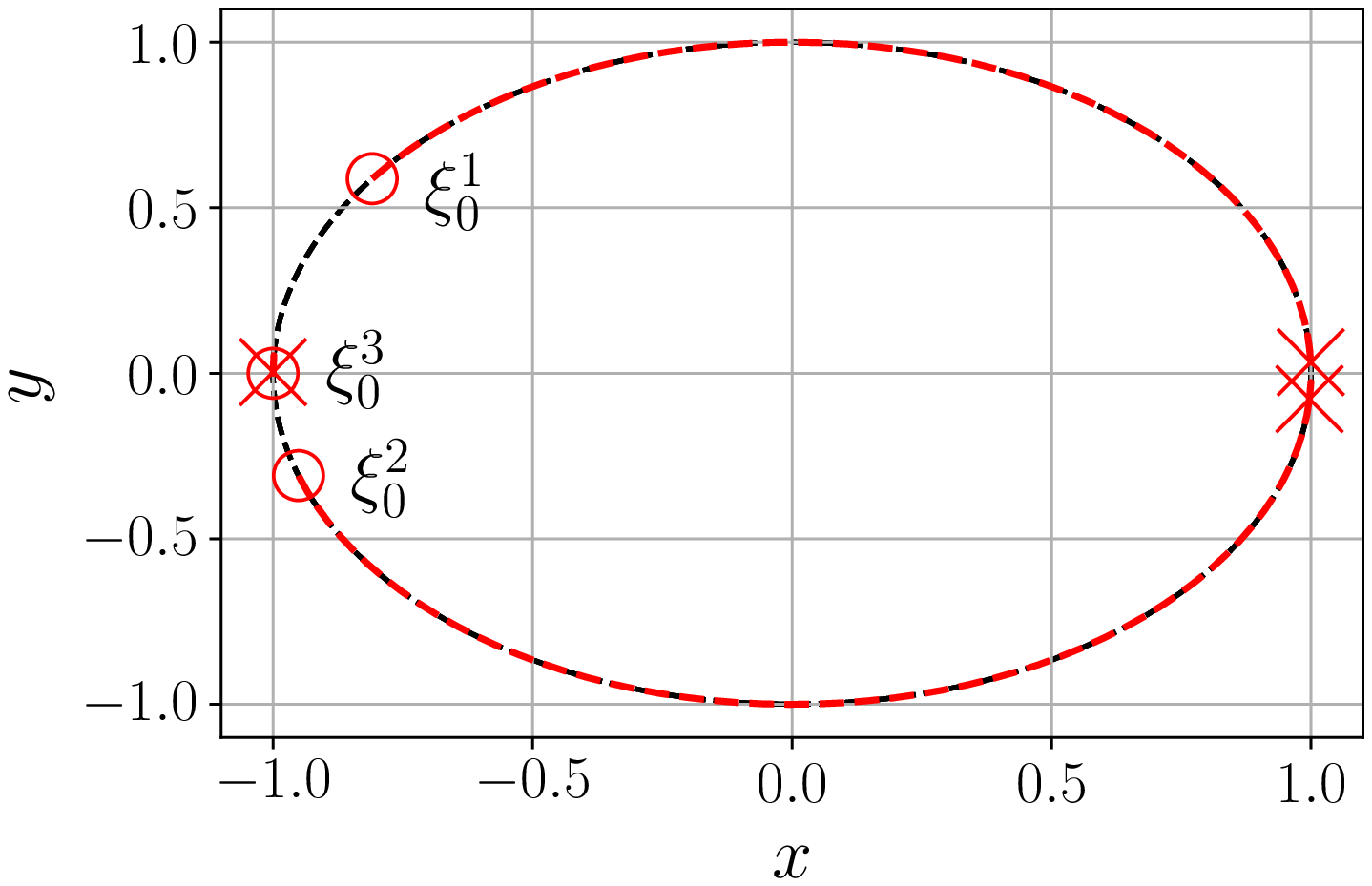}
\end{subfigure}
    \caption{(Left) Visualization of the found PPO policy. The magnitude of the control input $u$ is represented by the colors shown in the colorbar. (Right) Simulations of the control policy for initial conditions $\xi_0^1$, $\xi_0^2$, and $\xi_0^3$ in the presence of small measurement noise of magnitude $ 0.1$. The red~`$\circ$' represents the initial state and the red~`$\times$' the state reached after 4 seconds of simulation.}
    \label{fig:policymapandsimulation_UC}
\end{figure}
\end{example}

\begin{example}[Discrete action space problem]\label{sec:ObstacleAvoidanceProblem}
An obstacle avoidance problem, similar to the problem discussed in~\cite{obstacleavoidance}, is considered. A vehicle has to move to a specific location, but to do so it has to steer past an obstacle. The dynamics of the vehicle are taken to be simple because the focus is on the obstacle avoidance task and not on the control of systems with complex dynamics. The dynamics of the vehicle are given by 
\begin{equation}
  \dot{\xi} = f(u): = \left[\begin{array}{c}
         1 \\
         u
    \end{array}{}\right] \quad \xi = \left[ \begin{array}{c}
         x  \\
         y 
    \end{array}{}\right] \in \pazocal{S},
    \label{eq:DynCar}
\end{equation}
where $u~\in~[-1,1]$ is the control input, $x$ is the horizontal position of the vehicle, $y$ is the vertical position of the vehicle, and $\pazocal{S} :=\left\{\xi\in\mathbb{R}^2 \mid x\in [0, 3], y \in \left[-1.5,1.5\right] \right\}$. The control input is discretized and given by $u \in \{-1,-0.5,0,0.5,1\}$. The goal of the vehicle is to steer either left or right past the obstacle in order to avoid it and reach the set-point $\xi^* = [x^*$~$y^*]^\top = [3$~$0]^\top$. An overview of the environment is shown in Figure~\ref{fig:policymapandsimulation_OA}. The reward function used has a global maximum at the set-point and is symmetric in the sense that $R(x,y) = R(x,-y)$. The control policy $\pi^*(o(\xi))$ found numerically is visualized in the left figure in Figure~\ref{fig:policymapandsimulation_OA}.
\begin{figure}[]
    \centering
\begin{subfigure}[t]{.5\linewidth}
    \centering
    \includegraphics[width = 1\linewidth, height=3.25cm]{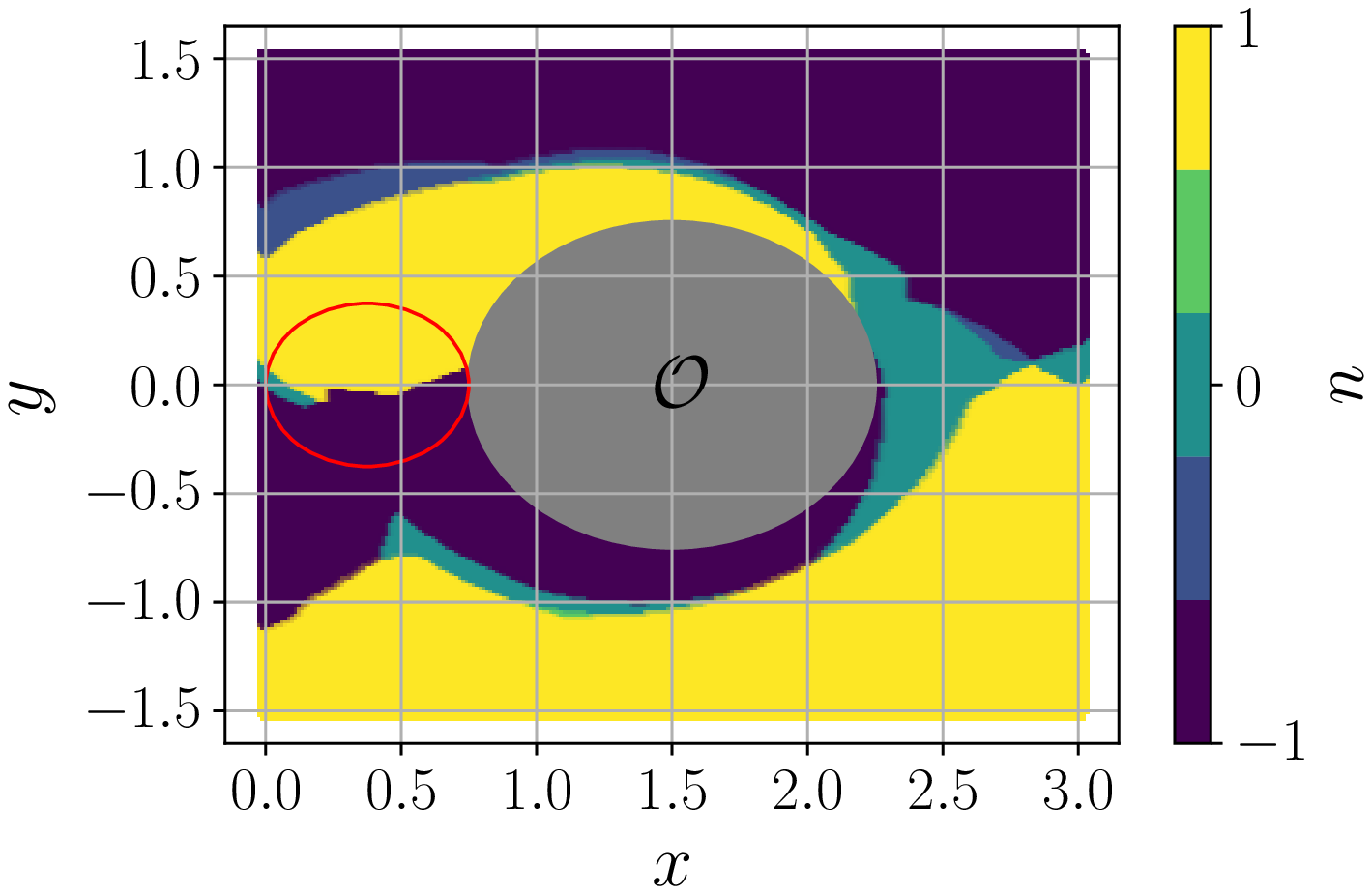}
\end{subfigure}%
\begin{subfigure}[t]{.5\linewidth}
    \centering
    \includegraphics[width = 1\linewidth, height=3.25cm]{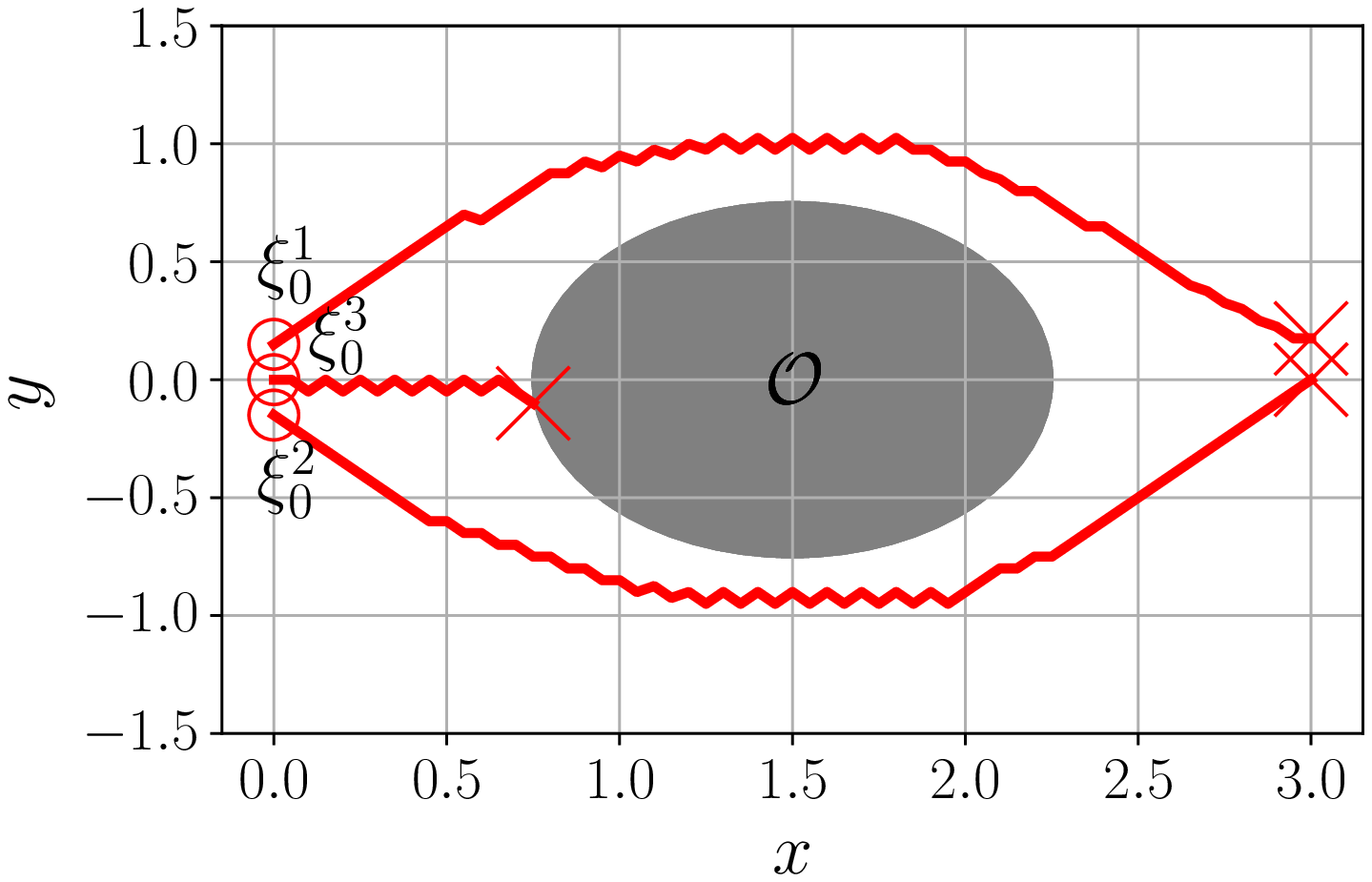}
\end{subfigure}
    \caption{(Left) Visualization of the found DQN policy. The magnitude of the control input $u$ is represented by the colors shown in the colorbar. In red, the critical points are marked. (Right) Simulations of the control policy for initial conditions $\xi_0^1$, $\xi_0^2$, and $\xi_0^3$ in the presence of small measurement noise of magnitude $ 0.1$. 
    The obstacle is denoted by~$\pazocal{O}$, the initial state by the red~`$\circ$', and the terminal state by the red~`$\times$'.}
    \label{fig:policymapandsimulation_OA}
\end{figure}
In the left figure in Figure~\ref{fig:policymapandsimulation_OA}, it can be seen that the policy, roughly speaking, steers the vehicle left past the obstacle when $y>0$, and right past the obstacle when $y < 0$. In the right figure in Figure~\ref{fig:policymapandsimulation_OA}, the system is initialized at $\xi_0^1 = [0$~$0.15]^\top$, $\xi_0^2 = [0$~$0]^\top$, and $\xi_0^3 = [0$~$-0.15]^\top$. For this simulation, a small amount of noise of magnitude $0.1$ is added to the measurement of the state $y$. In the right figure in Figure~\ref{fig:policymapandsimulation_OA}, it can be seen that the solution starting from $y=0.15$ steers left past the obstacle, and the solution starting from $y=-0.15$ steers right past the obstacle. However, the solution starting from $y=0$ crashes into the obstacle in the presence of small measurement noise. The found policy is hence not robust against small measurement noise, meaning that for a small amount of noise the system can crash into the obstacle, as can be observed in the right figure in Figure~\ref{fig:policymapandsimulation_OA}. The area around the point~$\xi_c= [0.4$~$0]^\top$ is considered a set of critical points for this problem.
\end{example}

\section{Hysteresis-Based RL}\label{sec:HybridDQN}
The examples in Section~\ref{sec:motivation}, show that the control policies derived from PPO and DQN may not be robust against small measurement noise. To guarantee robustness to small measurement noise, an algorithm is proposed that uses logic with an existing RL algorithm.
We refer to it as HyRL and the outcome is a hybrid closed-loop system that is robust against small measurement noise. In this section, we propose the HyRL algorithm and illustrate it in the unit circle example. 
\subsection{Outline}\label{sec:hybridDQNOutline}
To achieve robustness against measurement noise of a given magnitude, an algorithm needs to be designed to prevent solutions from getting stuck around the critical points. The following conceptual procedure is proposed: 
\begin{itemize}
    \item The environment is split up into two overlapping sets where the intersection is the set of critical points. 
    \item The overlapping sets are extended by means of backwards propagating the system around the set of critical points.
    \item The RL method of choice is used to find two new control policies for each of the extended overlapping sets.
    \item A hybrid system is build that incorporates a hysteresis switching effect between the two newly found control policies, thereby inducing robustness against measurement noise with the given magnitude.
\end{itemize}
\subsection{Class of systems}\label{sec:classofsystems}
The focus is on systems where optimal control policies generate trajectories evolving in opposite directions from some region of the state space for a small change in the state. These systems arise in the context of RL when the environmental rewards are symmetric.
Environmental rewards are symmetric with respect to some hyperplane $H$ if for states $x, \chi \in \pazocal{S}$, where $x \not=\chi$ and $\pazocal{S}$ is the set of all states in the environment, that are positioned on opposing sides of the hyperplane $H$ and have equal distance to the hyperplane $H$, the vector spanned from $x$ to $\chi$ is orthogonal to the hyperplane $H$ and the reward function $R:\pazocal{S}\rightarrow\mathbb{R}$ is a non-injective surjective function, i.e., $R(x)=R(\chi)$.
Then, two trajectories evolving in opposite directions can be found from some region of the state space. For instance, in Example~\ref{sec:unitcircleDQN}, environmental rewards are symmetric with respect to the hyperplane $H:=\left\{\xi\in\pazocal{S} \mid y=0 \right\}$. From the initial condition $\xi_0= [-1$~$0]^\top$, two unique optimal trajectories exist, namely a clockwise and a counterclockwise trajectory to the set-point $\xi^*$. Both trajectories obtain equal rewards, however evolve in opposite directions from the initial condition. Due to the symmetry of the environmental rewards, trajectories evolving in opposite directions are indeed found.

With the use of the set of critical points, the state space can be partitioned into two halves by imposing that trajectories do not leave the partition they start in. A set of critical points $\pazocal{M}^* \subset \pazocal{S}$ exists for a closed-loop system $\dot{\xi}=f(\xi,\pi(\xi))$ when the following property holds: 
\begin{itemize}
    \item[($\star$)] there exists $\rho>0$ such that for each state $\xi\in \pazocal{M}^*$ there exist initial states $z_0, z_1 \in \{\xi \} +\rho \mathbb{B}$ such that solutions $\xi_0, \xi_1$ to $\dot{\xi} = f(\xi,\pi(\xi))$ starting from $z_0, z_1$, respectively, satisfy 
    \begin{equation}
    \xi_0(t) \in \pazocal{M}_0 \text{ and } \xi_1(t) \in \pazocal{M}_1 \text{ for all } t\geq 0,
    \end{equation}
    where $\pazocal{M}_0$ and $\pazocal{M}_1$ are partitions of the environment $\pazocal{S}$ with properties $\pazocal{M}_0 \cup \pazocal{M}_1 = \pazocal{S}$ and $\pazocal{M}_0 \cap \pazocal{M}_1 = \pazocal{M}^*$.
\end{itemize}
For instance, in Example~\ref{sec:unitcircleDQN}, we have shown that the policy, roughly speaking, steers solutions clockwise when in the top half of the circle and counterclockwise when in the bottom half of the circle. Hence, ($\star$) holds. We denote the critical point for which solutions evolve in opposite directions from as $\xi_c$. For Example~\ref{sec:unitcircleDQN}, the critical point is given by $\angle\xi_c = \pi$, where $\angle \xi$ denotes the angle of $\xi$ in radians, as can be seen in the left figure in Figure~\ref{fig:policymapandsimulation_UC}. The set of critical points is defined as $\pazocal{M}^* := \{\xi \in \pazocal{S}^1\mid \angle\xi = \angle \xi_c\}$. The state space is then partitioned in two halves: the set of states where solution travel clockwise and counterclockwise given by $\pazocal{M}_0:=\{\xi \in \pazocal{S}^1 \mid \angle \xi \in [0, \angle \xi_c]\}$ and $\pazocal{M}_1:=\{\xi \in \pazocal{S}^1 \mid \angle\xi \in[\angle \xi_c,2\pi)\}$, respectively. For Example~\ref{sec:ObstacleAvoidanceProblem}, we have shown that the policy, roughly speaking, steers solutions left past the obstacle when $y$ is positive and right past the obstacle when $y$ is negative. Due to this, it can be shown that ($\star$) holds for Example~\ref{sec:ObstacleAvoidanceProblem} as well.
\subsection{Proposed HyRL algorithm}\label{sec:algorithmsteps}
An overview of the steps of the design of the HyRL algorithm is given below. For illustration purposes, the steps are elaborated for Example~\ref{sec:unitcircleDQN}. 
\begin{steps}[align=left, leftmargin=0pt, labelindent=\parindent, listparindent=\parindent, labelwidth=0pt, itemindent=!]
    \item Run an RL algorithm of choice on all states $\xi$ in the environment $\pazocal{S}$ to find a control policy $\pi^*(o(\xi))$. 
    \begin{adjustwidth}{\parindent}{0pt}
    \textit{For Example~\ref{sec:unitcircleDQN}, PPO is used to find a control policy; the policy is visualized in the left figure in Figure~\ref{fig:policymapandsimulation_UC}.}
    \end{adjustwidth}
    \item Determine the set of critical points $\pazocal{M}^*$ present in the control policy $\pi^*(o(\xi))$.
    \begin{adjustwidth}{\parindent}{0pt}
    \indent \textit{For the examples in this paper, an algorithm is used to automatically find $\pazocal{M}^*$. More details on this algorithm can be found on the aforementioned GitHubs.}
    \end{adjustwidth}
    \item Define two new sets of states $\pazocal{M}_0$ and $\pazocal{M}_1$, where $\pazocal{M}_0 \cup \pazocal{M}_1 = \pazocal{S}$ and $\pazocal{M}_0 \cap \pazocal{M}_1 = \pazocal{M}^*$. 
    \begin{adjustwidth}{\parindent}{0pt}
    \indent \textit{For Example~\ref{sec:unitcircleDQN}, the partitions are given by $\pazocal{M}_0:=\{\xi \in \pazocal{S}^1 \mid \angle \xi \in [0, \angle \xi_c]\}$ and $\pazocal{M}_1:=\{\xi \in \pazocal{S}^1 \mid \angle\xi \in[\angle \xi_c,2\pi)\}$.}
    \end{adjustwidth}
    \item Define two control policies, $\pi_0(o(\xi))$ and $\pi_1(o(\xi))$, in which $\pi_i(o(\xi)):=\pi^*(o(\xi))$ for $\xi \in \pazocal{M}_i$ and each $i \in \{0,1\}$. 
   \begin{adjustwidth}{\parindent}{0pt}
    \indent \textit{For Example~\ref{sec:unitcircleDQN}, the policy $\pi_0(o(\xi)):=\pi^*(o(\xi))$ for $\xi \in \pazocal{M}_0$ moves the system clockwise and policy $\pi_1(o(\xi)):=\pi^*(o(\xi))$ for $\xi \in \pazocal{M}_1$ moves the system counterclockwise.}
    \end{adjustwidth}
    \item For each $i\in \{0,1\}$, backward propagate the closed-loop system for a horizon $T \in \mathbb{R}_{>0}$ in the neighborhood of the critical points by using $\pi_i(o(\xi))$ for $\xi \in \pazocal{M}_i$. The states reached outside the domain $\pazocal{M}_i$ are stored in the set $\pazocal{X}_i$. The extended set is then defined as $\pazocal{M}^\text{ext}_i:=\pazocal{M}_i \cup \pazocal{X}_i$.
    \begin{adjustwidth}{\parindent}{0pt}
    \indent \textit{For Example~\ref{sec:unitcircleDQN}, the two control policies render the set $\pazocal{M}^*$ stable for the backward in-time system, meaning that during backward propagation, the system moves toward the critical point. The backward in-time system is obtained by switching the sign of the dynamics of the system. The backward in-time system of (\ref{eq:UnitcircleDyn}) is $\dot{\xi}=-f(\xi,u)$. A horizon time $T$ of $0.5$ seconds is used. The resulting sets $\pazocal{M}^\text{ext}_0$ and $\pazocal{M}^\text{ext}_1$ are given by $\pazocal{M}^\text{ext}_0 := \{\xi \in \pazocal{S}^1 \mid \angle \xi \in [0, 1.13\pi]\}$ and $\pazocal{M}^\text{ext}_1 := \{\xi \in \pazocal{S}^1 \mid \angle \xi \in [0.87\pi, 2\pi)\}$.}
    \end{adjustwidth}
    \item For each $i\in \{0,1\}$, run the same RL algorithm for $\pazocal{M}^\text{ext}_i$ to find policies $\pi_i^*(o(\xi))$ for each $\xi \in \pazocal{M}^\text{ext}_i$.
 \begin{adjustwidth}{\parindent}{0pt}
    \indent \textit{For Example~\ref{sec:unitcircleDQN}, the PPO algorithm is run for each of the two new environments $\pazocal{M}^\text{ext}_0$ and $\pazocal{M}^\text{ext}_1$, to find policies $\pi_0^*(o(\xi))$ for $\xi \in \pazocal{M}^\text{ext}_0$ and $\pi_1^*(o(\xi))$ for $\xi \in \pazocal{M}^\text{ext}_1$. $\pi_0^*(o(\xi))$ moves the system clockwise and $\pi_1^*(o(\xi))$ moves the system counterclockwise.}
    \end{adjustwidth}
    \item Build the hybrid system~$\pazocal{H}=(C,F,D,G)$ with state $z = (\xi,q)\in \pazocal{S}^1\times \{0,1\}$, as discussed in Section~\ref{sec:hybridsystems}. The hybrid system is given by
\begin{equation}
     \left[\begin{array}{c}
        \dot{\xi}  \\
        \dot{q} 
   \end{array} \right]= F(z):= \left[\begin{array}{c}
          f(\xi,\pi_q^*(o(\xi))) \\
          0 
    \end{array}\right]  \quad  z \in C,
    \label{eq:hybridFLOW}
\end{equation}
\begin{equation}
    \left[\begin{array}{c}
        {\xi}^+  \\
        {q}^+ 
   \end{array} \right] = G(z):= \left[\begin{array}{c}
          \xi \\
          \left\{\begin{array}{cc}
              1 & \text{if} \quad q = 0 \\
              0 & \text{if} \quad q = 1
          \end{array}\right.
    \end{array}\right]  \quad  z \in D,
    \label{eq:HybridJUMP}
\end{equation}
where $C:=  \bigcup\limits_{q \in \{0,1\}} \left( \pazocal{M}^\text{ext}_q \times \{q\}\right)$ and $D:= \bigcup\limits_{q \in \{0,1\}} \left(\overline{(\pazocal{S} \setminus \pazocal{M}^\text{ext}_q)} \times \{q\}\right)$.
\begin{adjustwidth}{\parindent}{0pt}
\indent \textit{For Example~\ref{sec:unitcircleDQN}, the results of the hybrid system are discussed in Section~\ref{sec:hybridsims}.}
\end{adjustwidth}
\end{steps}
The hybrid system $\pazocal{H}$, given by (\ref{eq:hybridFLOW}) and (\ref{eq:HybridJUMP}), flows when in the set $\pazocal{M}^\text{ext}_i$ with the corresponding logic variable $q$, and jumps when this condition no longer holds. At each jump, the logic variable jumps from 1 to 0 or vice versa. By construction of this hybrid system, a hysteresis effect is implemented because the sets $\pazocal{M}^\text{ext}_0$ and $\pazocal{M}^\text{ext}_1$ overlap over the critical points $\pazocal{M}^*$. The system can no longer switch between opposing control inputs near $\pazocal{M}^*$, because the system has to flow a certain distance away from $\pazocal{M}^*$ before a jump (switch in control policy) can occur. The minimal width of the overlapping region has to be greater than the expected magnitude of noise on the measurement of the system. Then, the resulting control policy is robust against the expected measurement noise.

The computational cost of Step~6 is equal to or less than the cost of training the original policy (Step~1) for two reasons. Firstly, the training done in Step 6 starts with the agent found in Step 1. The new policies can be obtained by slight tweaks to the original agent. Secondly, the RL problem is simplified by splitting up the environment. The agent in Step~1 had to learn discontinuous behavior, that is, opposing control inputs for a small change in the state. This training is computationally challenging because the underlying parameterization is generally continuous.
However, in Step~6, two agents are used to realize the opposing control inputs. Therefore, each agent only learns one part of the opposing control inputs. Hence within one agent, no discontinuous behavior has to be learned. For example, in Example~\ref{sec:unitcircleDQN}, one agent moves the system clockwise while the other agent moves it counterclockwise.
\section{Examples}\label{sec:hybridsims}
HyRL, as introduced in Section~\ref{sec:algorithmsteps}, is applied to the unit circle and obstacle avoidance problems in Examples~\ref{sec:unitcircleDQN} and~\ref{sec:ObstacleAvoidanceProblem}, and the results are compared to those of PPO and DQN to show the added value of the HyRL algorithm. 
\begin{example}[Continuous action space problem revisited]
The steps of HyRL for the unit circle problem are discussed in Section~\ref{sec:algorithmsteps}. Measurement noise of magnitude $\epsilon =  0.1$ is applied to the angle measurement of the system. The resulting trajectories of HyRL and PPO for initial conditions $\angle\xi_0\in \left\{0.75\pi, 0.9\pi, \pi, 1.1\pi, 1.25\pi \right\}$ are shown in Figure~\ref{fig:Hybrid_DQN_UC}. In Figure~\ref{fig:hybrid_dqn_UC_q0}, the hybrid system is initialized with $q_0 = 0$. In Figure~\ref{fig:hybrid_dqn_UC_q1}, the hybrid system is initialized with $q_0=1$. For these simulations, HyRL and PPO are applied in the presence of the same measurement noise signal.
In Figure~\ref{fig:Hybrid_DQN_UC}, it can be seen that for the initial conditions $\angle\xi = \pi$, i.e., $\xi = [-1$~$0]^\top$, the PPO policy gets stuck at the critical point. However, it can be seen that for the same initial conditions and measurement noise signals, HyRL moves the system away from the critical point and stabilizes the system around the set-point. Notice that the solutions for initial conditions $\angle\xi_0= 0.9\pi$ and $\angle\xi_0= 1.1\pi$ move clockwise/counterclockwise depending on the initial value of the logic variable~$q$. Therefore, the width of the overlapping region is larger than $0.2\pi$ and also larger than the magnitude of the noise on the measurement of the system, i.e., $0.2\pi>\epsilon$, and thus the system obtained by HyRL is robust against measurement noise of magnitude $\epsilon$.
\begin{figure}[]
    \centering
\begin{subfigure}{.5\linewidth}
    \centering
    \includegraphics[width = 1\linewidth, height=3.25cm]{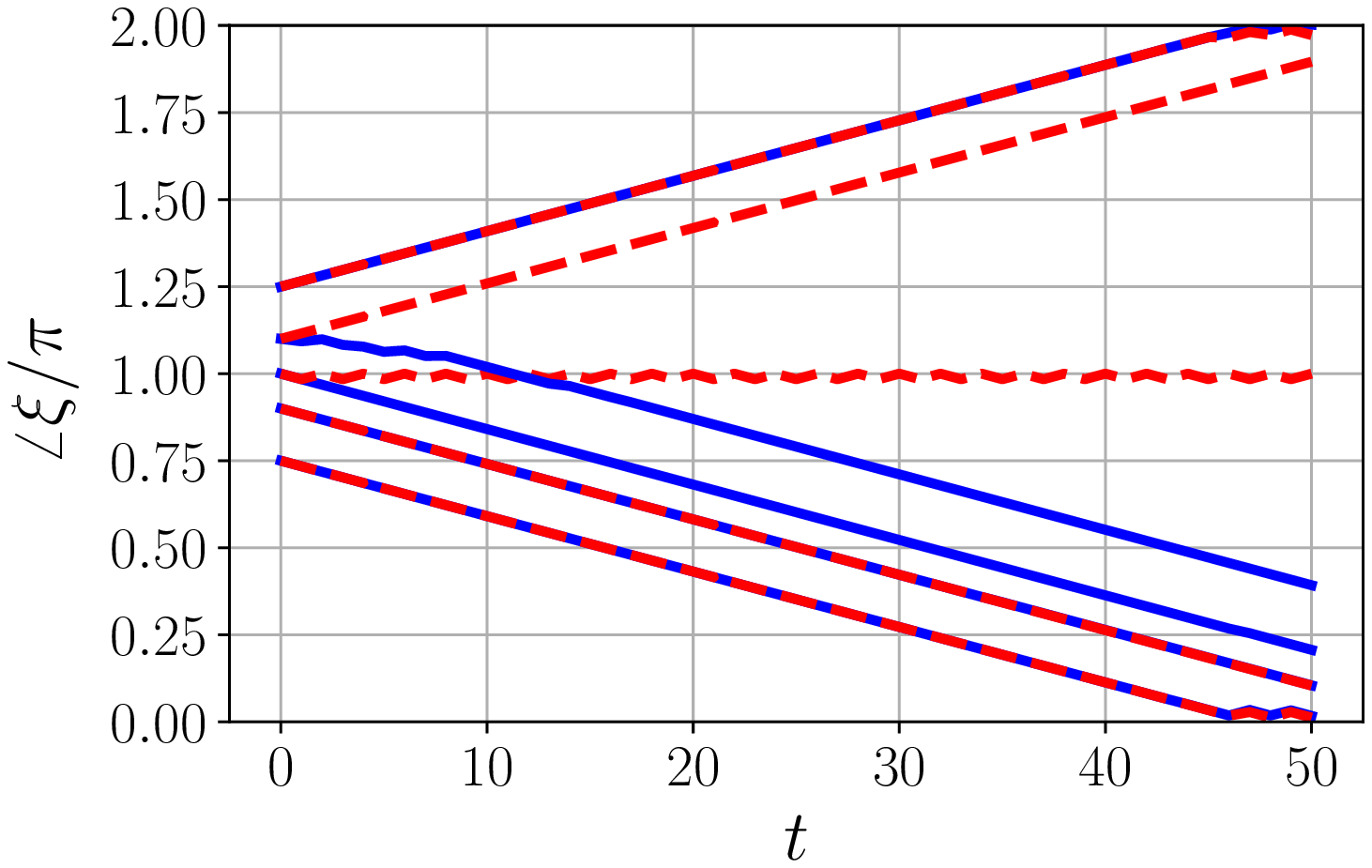}
    \caption{Initialized with $q_0 = 0$.}
    \label{fig:hybrid_dqn_UC_q0}
\end{subfigure}%
\begin{subfigure}{.5\linewidth}
    \centering
    \includegraphics[width = 1\linewidth, height=3.25cm]{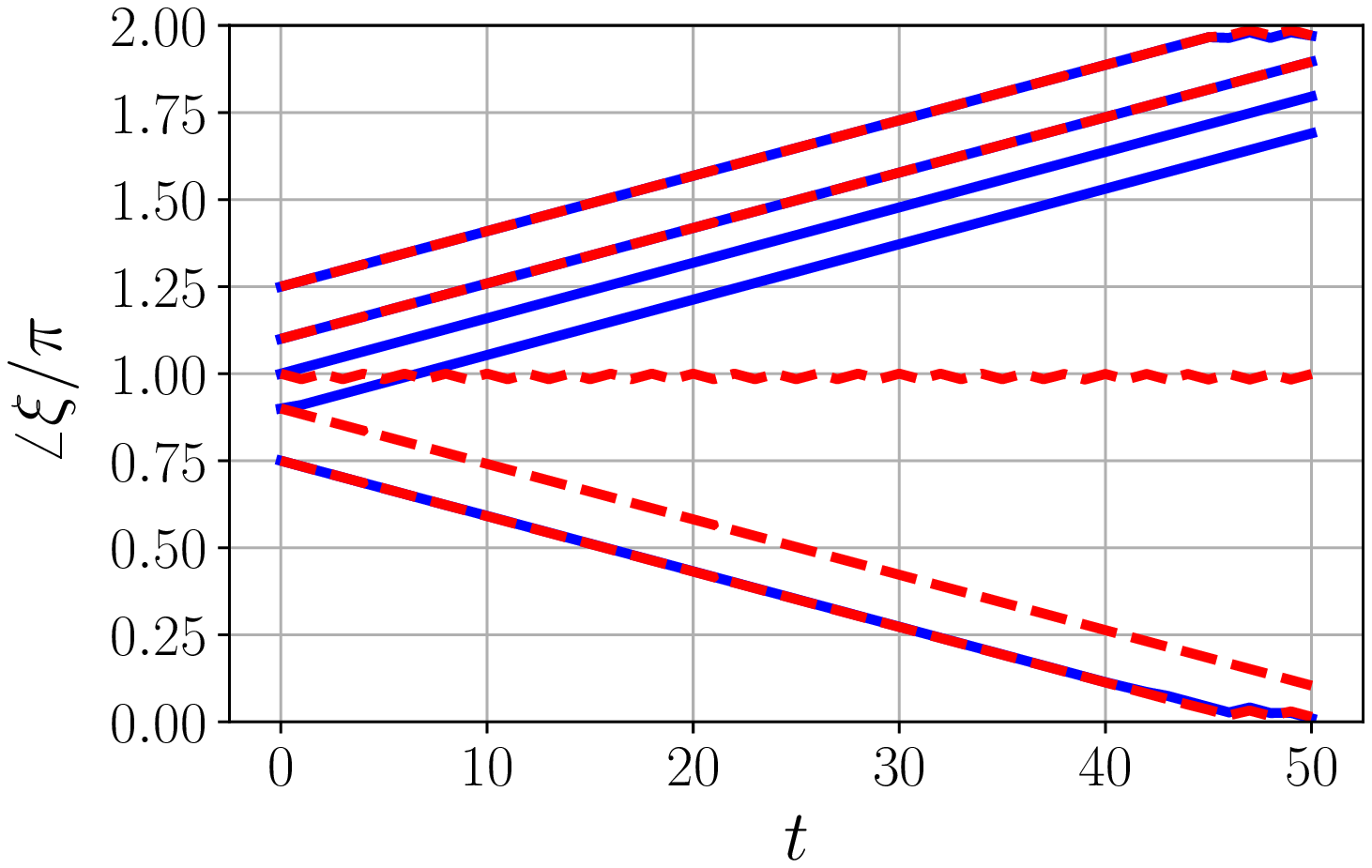}
    \caption{Initialized with $q_0 = 1$.}
    \label{fig:hybrid_dqn_UC_q1}
\end{subfigure}
    \caption{HyRL and PPO applied to the unit circle problem for initial conditions $\angle\xi_0\in \left\{0.75\pi, 0.9\pi, \pi, 1.1\pi, 1.25\pi \right\}$ in the presence of measurement noise $\epsilon = 0.1$. HyRL is shown by the blue lines and the PPO by the red lines.}
    \label{fig:Hybrid_DQN_UC}
\end{figure}
\end{example}
\begin{example}[Discrete action space problem revisited]
Measurement noise of magnitude $\epsilon =  0.1$ is applied to the measurement of the state~$y$. The resulting trajectories of HyRL and DQN for initial conditions $\xi_0 \in \left\{[0,-0.15], [0,-0.055], [0,0], [0,0.055], [0,0.15]\right\}$ are shown in Figure~\ref{fig:Hybrid_DQN_OA}. In Figure~\ref{fig:hybrid_dqn_OA_q0}, the hybrid system is initialized with $q_0 = 0$. In Figure~\ref{fig:hybrid_dqn_OA_q1}, the hybrid system is initialized with $q_0=1$. For these simulations, HyRL and DQN are applied in the presence of the same measurement noise signal.
In Figure~\ref{fig:Hybrid_DQN_OA}, it can be seen that for the trajectory near $\xi = [0.4$~$0]^\top$, the DQN policy crashes into the obstacle. However, it can be seen that for the same initial conditions and measurement noise signals, HyRL steers the system past the obstacle and to the set-point. Notice that solutions for initial conditions $\xi_0= [0$~$-0.055]^\top$ and $\xi_0= [0$~$0.055]^\top$ steer left/right past the obstacle depending on the initial value of the logic variable~$q$. Therefore, the width of the overlapping region is larger or equal to $0.11$ and also larger than the magnitude of the noise on the measurement of the system, i.e., $0.11 > \epsilon$, and thus the system obtained by HyRL is robust against measurement noise of magnitude $\epsilon$.
\begin{figure}[]
    \centering
\begin{subfigure}{.5\linewidth}
    \centering
    \includegraphics[width = 1\linewidth, height=3.25cm]{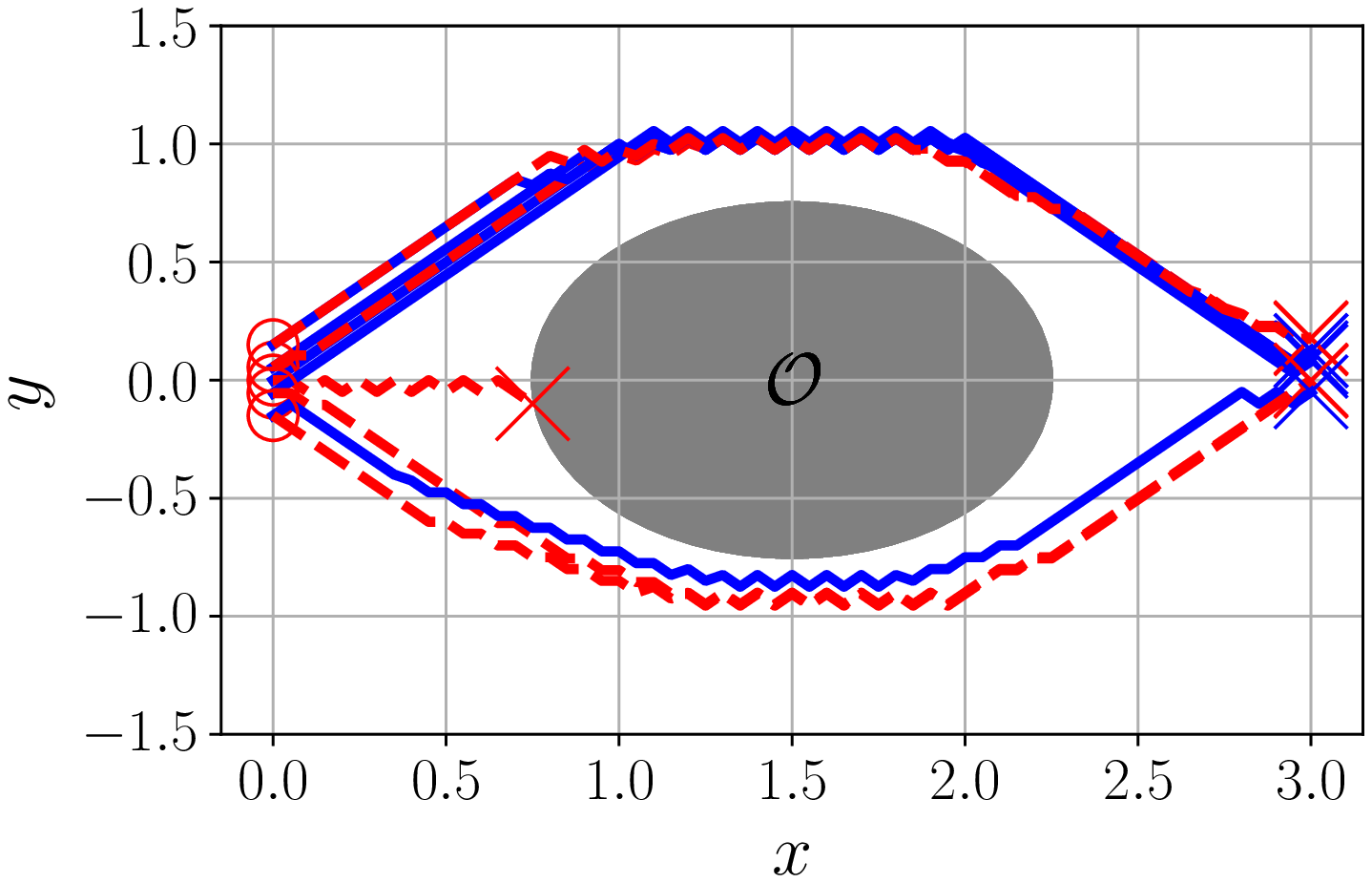}
    \caption{Initialized with $q_0 = 0$.}
    \label{fig:hybrid_dqn_OA_q0}
\end{subfigure}%
\begin{subfigure}{.5\linewidth}
    \centering
    \includegraphics[width = 1\linewidth, height=3.25cm]{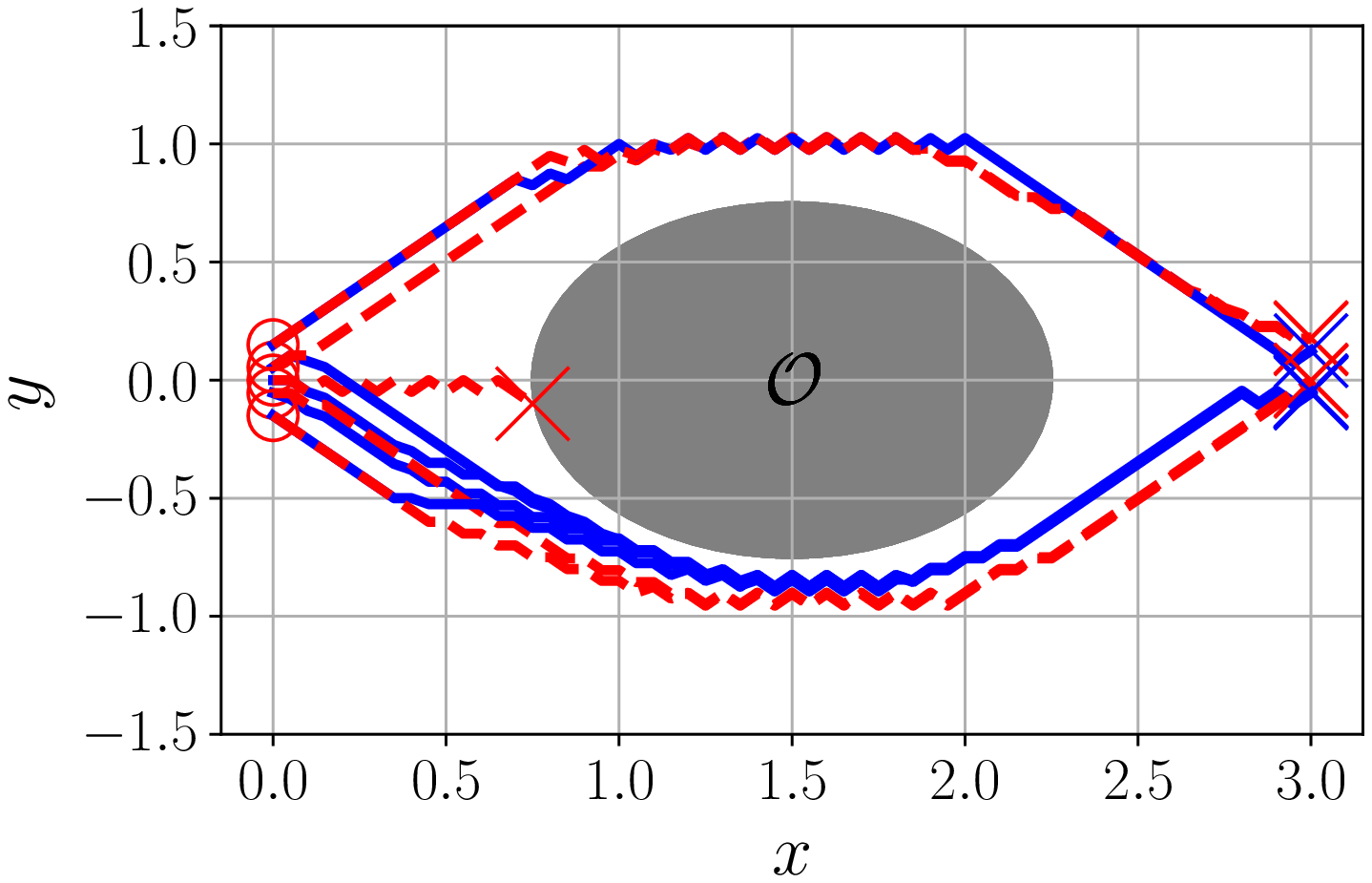}
    \caption{Initialized with $q_0 = 1$.}
    \label{fig:hybrid_dqn_OA_q1}
\end{subfigure}
    \caption{HyRL and DQN applied to the obstacle avoidance problem for initial conditions $\xi_0 \in \left\{[0,-0.15], [0,-0.055], [0,0], [0,0.055], [0,0.15]\right\}$ in the presence of measurement noise $\epsilon = 0.1$. HyRL is shown by the blue lines and the DQN by the red lines. The obstacle is denoted by~$\pazocal{O}$, the initial state by the red~`$\circ$', and the terminal state by the red~`$\times$'.}
    \label{fig:Hybrid_DQN_OA}
\end{figure}
\end{example}

\section{CONCLUSION}
In this paper, the problem related to critical points for RL-based control policies is discussed. We have shown with the use of two example control problems that the derived policies from PPO and DQN lack robustness guarantees. Motivated by these issues, we proposed a new algorithm, named HyRL. HyRL gives rise to a hybrid closed-loop system that is robust against small measurement noise. We have demonstrated its stability and robustness properties in two examples for which PPO and DQN fail. With HyRL, the robustness of other existing RL methods can be improved by making them hybrid, and this is part of future research.%

\addtolength{\textheight}{-12cm}   

\bibliographystyle{IEEEtran}
\bibliography{references}

\end{document}